\documentclass[11pt,a4paper]{article}
\usepackage[]{acl2020}
\usepackage{times}
\usepackage{latexsym}
\usepackage{amsmath}
\usepackage{courier}
\usepackage{graphicx}
\usepackage{mathrsfs}
\usepackage{amsfonts}
\usepackage{subfigure}
\usepackage{multirow}
\usepackage{url}
\usepackage{comment}
\usepackage{algorithm}
\usepackage{algorithmic}

\author{
Wangchunshu Zhou\thanks{\ \ Equal contribution.} ~\thanks{\ \ Corresponding author} ~~~ Qifei Li$^{*}$ ~~~ Chenle Li\\
Beihang University, Beijing, China\\
{\tt zhouwangchunshu@buaa.edu.cn}}
\date{}

\date{}
 
\begin{document}
%
\title{Learning to Predict Persona Information for \\ Dialogue Personalization without Explicit Persona Description}
\maketitle
\begin{abstract}
Personalizing dialogue agents is important for dialogue systems to generate more specific, consistent, and engaging responses. However, most current dialogue personalization approaches rely on explicit persona descriptions during inference, which severely restricts its application. In this paper, we propose a novel  approach that learns to predict persona information based on the dialogue history to personalize the dialogue agent without relying on any explicit persona descriptions during inference. Experimental results on the PersonaChat dataset show that the proposed method can improve the consistency of generated responses when conditioning on the predicted profile of the dialogue agent (i.e. ``self persona''), and improve the engagingness of the generated responses when conditioning on the predicted persona of the dialogue partner (i.e. ``their persona''). We also find that a trained persona prediction model can be successfully transferred to other datasets and help generate more relevant responses.
\end{abstract}


\section{Introduction}

Recently, end-to-end dialogue response generation models~\cite{sordoni2015neural,serban2016building,bordes2016learning} based on recent advances of neural sequence-to-sequence learning models~\cite{sutskever2014sequence,vaswani2017attention} have gained increasing popularity as they can generate fluent responses. However, as the dialogue agent is trained with datasets containing dialogues from many different speakers, it can not generate personalized responses for the current speaker, making the generated responses less relevant and engaging~\cite{li2016persona}. 

To address this problem, recent studies attempt to personalize dialogue systems by generating dialogue responses conditioning on given persona descriptions have been shown to help dialogue agents perform better~\cite{zhang2018personalizing,mazare2018training}. However, a major drawback of the current dialogue agent personalization approaches is that they require explicit persona descriptions in both training and inference stages, which severely limits their application in real-world scenarios because detailed persona descriptions for current speakers are not available in most scenarios. Another problem is that current dialogue personalization approaches are not interpretable and the role of additional persona information is unclear.

In this paper, we propose a novel dialogue agent personalization approach that automatically infers the speaker's persona based on the dialogue history which implicitly contains persona information. Our model generates personalized dialogue responses based on the dialogue history and the inferred speaker persona, alleviating the necessity of the persona description during inference.

Specifically, we propose two different approaches to perform persona detection. The first approach learns a ``persona approximator'' which takes dialogue history as the input and is trained to approximate the output representation of a persona encoder that takes explicit persona description as the input. The second approach instead addresses the persona detection problem as a sequence-to-sequence learning problem and learns a ``persona generator'' which takes the dialogue history as the input and generates the persona description of the speaker. This approach provides a stronger supervision signal compared with the first approach and is more interpretable as the encoded persona information can be decoded to reconstruct the detected persona description. 

\begin{figure*}[htbp]
\centering
\subfigure{
\centering
\includegraphics[width=0.5\textwidth]{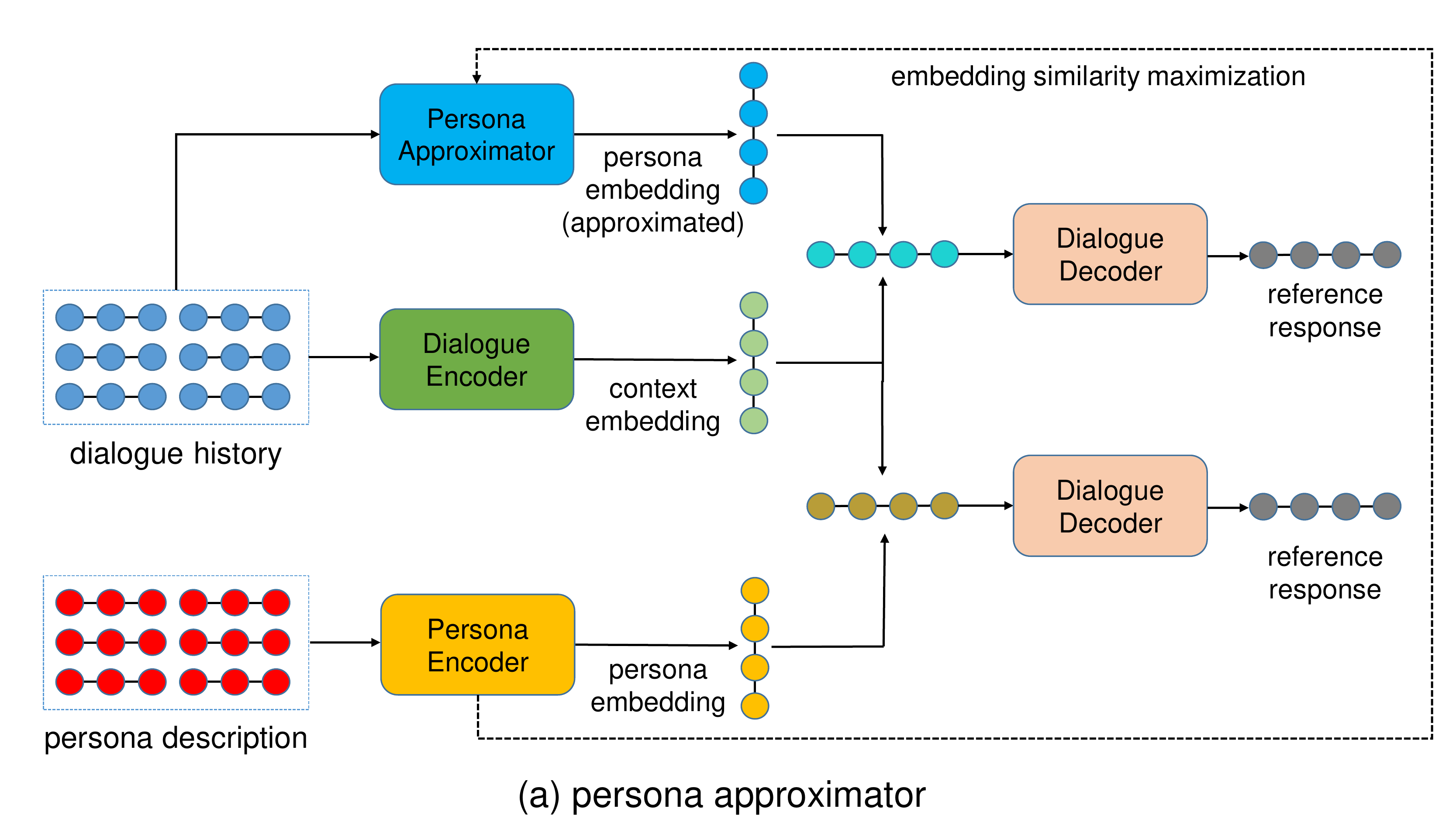}
}%
\subfigure{
\centering
\includegraphics[width=0.5\textwidth]{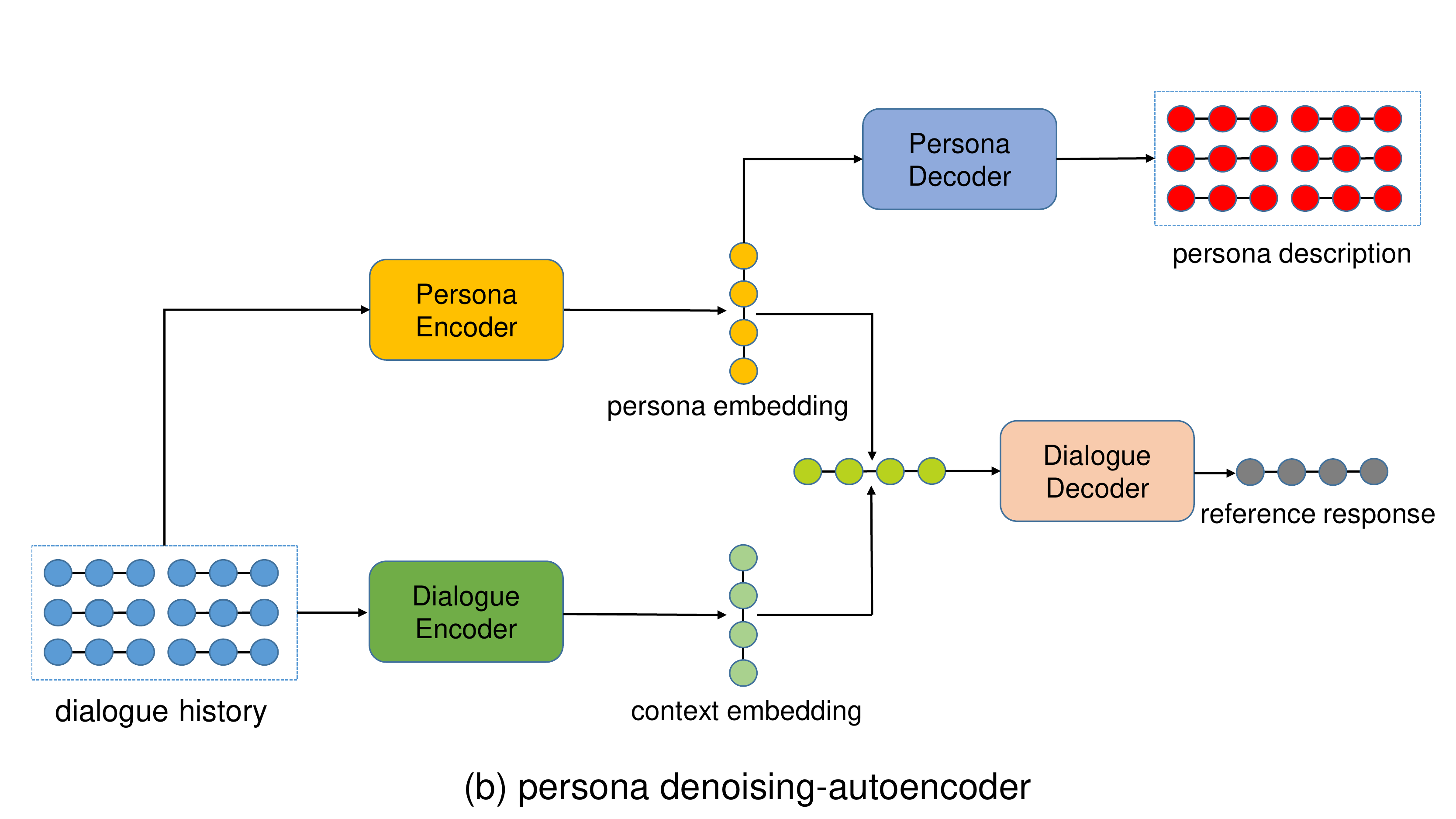}
}%
\centering
\caption{Illustration of the proposed persona detection models. The persona approximator is on the left. It is trained to maximize the embedding similarity between persona embedding approximated by the persona approximator and the persona encoder, which is obtained by taking dialogue history and persona description respectively. The persona generator is on the right, which is trained to recover persona description from the dialogue history, thus can also be viewed as a ``persona denosing-autoencoder''. The persona decoder is employed for training and only the persona encoder is used during inference.}
\label{model}
\end{figure*}

Our proposed approach can be used to incorporate both ``self-persona'' which is the persona information of the dialogue agent, and ``their-persona'' which is the persona information of the dialogue partner. On one hand, generating dialogue responses conditioning on the inferred ``self-persona'' can help the dialogue agent maintain a consistent persona during the conversation, thus enhancing the consistency of generated responses without the need of a pre-defined persona description for every dialogue agent. On the other hand, generating dialogue responses conditioning on the predicted persona of the dialogue partner helps the dialogue model generate more engaging responses that are relevant to its dialogue partner. The ability to automatically infer the persona information of the dialogue partner is particularly attractive because in many real-world application scenarios, the persona information of the user is hardly available before the dialogue starts.
In addition, to facilitate training and tackle the problem of lacking training data, we propose to train the persona detection model with multi-task learning by sharing layers and training jointly with the dialogue context encoder in both approaches. 

Our experiments on dialogue datasets with and without the persona description demonstrate the effectiveness of the proposed approach and show that a trained persona detection model can be successfully transferred to datasets without persona description.

\section{Related Work}



Preliminary study on dialogue personalization~\cite{li2016persona} attempts to use a persona-based neural conversation model to capture individual characteristics such as background information and speaking style.  However, it requires the current speaker during inference to have sufficient dialogue utterances included in the training set, which is quite restricted by the cold-start problem.

More recently, \citet{zhang2018personalizing} released the PersonaChat dataset which incorporates \textit{persona} of two speakers represented as multiple sentences of profile description to personalize dialogue agents. They propose a profile memory network by considering the dialogue history as input and then performing attention over the persona to be combined with the dialogue history. ~\citet{mazare2018training} proposed to train a persona encoder and combine the encoded persona embedding with context representation by concatenation. The combined representation is then fed into the dialogue decoder to generate personalized responses. ~\cite{yavuz2019deepcopy} designed the DeepCopy model, which leverages copy mechanism to incorporate persona texts and~\citet{madotto2019personalizing} propose to use meta-learning to adapt to the current speaker quickly, their approach also requires several dialogues of the speaker to perform dialogue personalization, which is different from our approach. ~\citet{welleck2018dialogue} propose a dialogue natural language inference dataset and use it to measure and improve the consistency of the dialogue system. More recently, ~\citet{zheng2019personalized} propose personalized dialogue generation with diversified traits.~\citet{song2020generate} introduce a multi-stage response generation stage to improve the personalization of generated responses. ~\citet{DBLP:conf/acl/WuLWCWFHW20} propose a variational response generator to better exploit persona information. Different from the aforementioned works, our approach does not require persona information during test time, which makes it more generally applicable.


\section{Methodology}


The motivation behind the proposed approach is that we can learn to detect the profile (i.e., persona) of dialogue speakers based on the dialogue history, which is demonstrated by experimental results in \citet{zhang2018personalizing} that we can train a model to effectively distinguish the corresponding persona from randomly sampled negative persona based on the dialogue history. 

The key idea is to jointly train a persona detection model with a conventional dialogue response generation model. The persona detection model is trained with persona description to infer the persona information based on the dialogue history, which provides persona information for the dialogue model, thus alleviating the necessity of provided persona information during test time. We propose two different persona detection models. The first model is a ``persona approximator'' and the second is a ``persona generator''. An overview of the proposed models is illustrated in Figure \ref{model}. We describe them in detail in this section, together with a multi-task learning objective which facilitates the training stage of the model.

\subsection{Task Definition}

Given a dialogue dataset $\mathcal{D}$ with personas, an example of the dataset can be represented as a triplet $(h, p, r)$. Specifically, $h = \{u_{1}, u_{2}, ..., u_{nh}\}$, which represents the dialogue history with $nh$ utterances. $p = \{p_{1}, p_{2}, ..., p_{np} \}$, which represents a persona with $np$ profile sentences. $r$ represents the ground-truth response. Existing personalized dialogue models learn a dialogue response generation model $G$ which takes $h$ and $p$ as input during inference and generates a personalized response $G(h,p)$. Our goal is to learn a persona detection model $D$ which enables the dialogue model to generate personalized response $G(h,D(h))$ without relying on given persona description $p$ during test time. In this way, the persona description in the dataset is used to train the personalized dialogue agent and after training, our model should be able to generate personalized dialogue responses without relying on persona description. 

\subsection{Persona Approximator}

The idea of persona approximator is that given a trained personalized dialogue model with persona encoder which takes the persona description as input and outputs the persona embedding, we can train a persona approximator which takes the dialogue history as input and learns to output a persona embedding which is similar with that encoded by the trained persona encoder. Persona embedding approximation is possible as dialogue history is shown to be sufficient for discriminating the corresponding persona~\cite{zhang2018personalizing}.

Formally, given dialogue history $h$ and persona description $p$, the persona encoder $E$ takes $p$ as input and outputs persona embedding $emb(p) = E(p)$. The proposed persona approximator $A$ takes $h$ as input and outputs the approximated persona embedding $a = A(h)$. The training objective of $A$ is to optimize the embedding similarity (e.g. cosine similarity) between $a$ and $emb(p)$.

We discuss several pros and cons of the proposed persona approximator here. The advantage of this approach is that it alleviates the requirement of persona description during training and can incorporate several off-the-shelf personalized dialogue models with persona encoder seamlessly. However, as the persona encoder itself is far from perfect and non-interpretable, a persona approximator which is trained to approximate the persona encoder may also be sub-optimal and even less interpretable. Another issue is that the persona approximator can only be trained after training the dialogue model and persona encoder. To alleviate this problem and train an interpretable persona detection model more effectively, we propose another persona detection model which is named ``persona generator''.

\subsection{Persona Generator}

As dialogue history can be used to predict the corresponding persona, which is demonstrated by~\citet{zhang2018personalizing}, we hypothesize that dialogue history implicitly contains the persona of dialogue partners. Therefore, we argue that a good persona detection model should be able to reconstruct the dialogue partners' persona descriptions based on the dialogue history. Based on this insight, we propose a ``persona generator'' model which formulates the persona detection problem as a sequence-to-sequence learning problem and train the persona generator to recover the textual persona description of dialogue partners from the dialogue history. 


Formally, the persona generator receives the dialogue history $h$ as input and is trained to generate the persona description $p$, which is a sequence of tokens $p_{i}$ of length $n$. The persona generator is trained by maximizing the likelihood of the ground-truth persona descriptions:
\begin{equation}\label{reward}
 \text{L}_{pg} = -\sum_{i = 1}^{n}{\log {P(p_{i}|p_{<i},h)}}
\end{equation}
As illustrated in Figure 1(b), the persona generator consists of a persona encoder and a persona decoder. During training, the persona encoder takes the dialogue history as input and outputs a persona embedding that represents the persona information of either the dialogue model or its dialogue partner. The persona embedding is then concatenated with the context embedding generated by the dialogue encoder and fed into the dialogue decoder to generate the response. In addition, the persona embedding is also fed into the persona decoder to generate the textual persona description of the dialogue partner. During inference, only the encoder of the trained persona generator will be used to provide persona information for the response generation model.

While previous dialogue personalization approaches, as well as the aforementioned persona approximator, generally train the persona encoder to maximize the likelihood of gold responses with MLE and can not ensure that the persona encoder actually captures useful persona information, the persona generator is directly trained to generate persona information from dialogue history, which enforces the persona information to be successfully captured. This approach also enhances the interpretability of the dialogue personalization procedure as the persona embedding encoded from dialogue history can be decoded into persona description with the decoder of trained persona generator.

\subsection{Multi-Task Learning}

Training the proposed persona detection models can be difficult because the available persona description is limited. To alleviate this problem, we propose to adopt multi-task learning~\cite{argyriou2007multi} by training the dialogue encoder jointly with the persona detection model. This is possible because both the dialogue encoder and the persona detection model take dialogue history as input and outputs a latent vector. The difference is that the dialogue context encoder is trained to provide direct information for response generation while the persona detection model is trained to predict persona description. These two tasks both require dialogue understanding and commonsense reasoning ability, which can be shared and help each other generalize better. We thus propose to adopt the multi-task learning paradigm to facilitate training. Specifically, we share the parameter of the first layer, which can be viewed as a general-purpose dialogue information encoder, between the dialogue context encoder and the persona detection model. 

In addition, we also train the persona detection model to maximize the likelihood of ground-truth responses together with the dialogue model, which ensures that the persona detection model not only encodes persona information but also helps generate more fluent dialogue responses. We control the relative importance between the original MLE objective and the training objectives of the proposed persona detection models by weighting the loss of persona detection objective with a hyperparameter $\alpha$ which is empirically set to 0.1 in our experiments.  

\section{Experiments}

\subsection{Dataset}

We conduct our experiments on PersonaChat dataset~\cite{zhang2018personalizing} which is a multi-turn chit-chat conversation dataset containing conversations between human annotators who are randomly assigned a “persona”. We experiment with two settings where the models are trained either with the persona description of themselves (i.e., self persona) or with the persona description of their dialogue partner (i.e., their persona). We present an example of the dataset in the Appendix.



In addition, we also expect our approach to be able to perform personalized dialogue response generation on other datasets (application scenarios) where persona description is not available even in the training set. Therefore, we also conduct experiments on the Dailydialog dataset~\cite{li2017dailydialog}, which is a multi-turn dialogue dataset in a similar domain with PersonaChat but without persona description, to explore the transferability of our approach.

\subsection{Evaluation Metrics}

For automated evaluation, we employ the following metrics following previous work:
\begin{itemize}
    \item \textbf{Perplexity} Following~\citet{zhang2018personalizing}, we use perplexity (ppl) to measure the fluency of responses. Lower perplexity means better fluency.
    \item \textbf{Distinct} Following~\cite{li2015diversity}, we calculate the token ratios of distinct bigrams (Distinct-2, abbreviated as Dst for convenience). We use this metric to measure the diversity of the responses.
    \item \textbf{Hits@1} Following~\citet{zhang2018personalizing}, Hit@1 measures the percentage of correct identification of a gold answer from a set of 19 distractors.
    \item \textbf{Consistency} In addition, we train a dialogue natural language inference model on the DNLI dataset~\cite{welleck2018dialogue} by fine-tuning BERT~\cite{devlin2018bert}. We are able to achieve a test set accuracy of 88.60\%, which is comparable to the best reported model~\cite{welleck2018dialogue} (88.20\% accuracy). The consistency metric (Cons) is then defined following~\cite{madotto2019personalizing}:
\begin{equation}
\small
\begin{array}{l}{\operatorname{NLI}\left(u, p_{j}\right)=\left\{\begin{array}{cc}{1} & {\text { if } u \text { entails } p_{j}} \\ {0} & {\text { if } u \text { is independent to } p_{j}} \\ {-1} & {\text { if } u \text { contradicts } p_{j}}\end{array}\right.} \\ \\
{Cons(u)=\sum_{j}^{m} \mathrm{NLI}\left(u, p_{j}\right)}\end{array}
\end{equation}
\end{itemize}


As automated metrics generally fail to correlates well with human evaluation~\cite{liu2016not,zhou2020learning}. We also systematically conduct human evaluation to further evaluate the proposed method. Specifically, we invite 20 human annotators that are all graduate students with good English proficiency to evaluate the quality of the model. Following~\citet{zhang2018personalizing}, we ask human annotators to interact with compared models and evaluate the fluency,
engagingness, and consistency of the model (scored between 1-
5). In addition, the degree of personalization of the model is measured by the ability of human annotators to detect the model’s profile after the conversation, which is measured by displaying the real persona description together with a randomly sampled persona description and asking the human annotator to select which is more likely to be the profile of the model. The persona detection metric is only available in PersonaChat where test persona is available.

\subsection{Compared Models}

\begin{table*}[t!]
\begin{center}
\scalebox{0.95}{
\begin{tabular}{lcccccccc}
\hline\hline
\multirow{2}{*}{\textbf{Method}} & \multicolumn{4}{c}{\bf Self Persona} & \multicolumn{4}{c}{\bf Their Persona}\\
 & \textbf{ppl} & \textbf{Dst}  & \textbf{Hits@1} &  \textbf{Cons} & \textbf{ppl} & \textbf{Dst} &  \textbf{Hits@1} & \textbf{Cons} \\ \hline
\bf TransferTransfo & 17.78 & \bf 21.5 & 80.1 & 0.32 & 18.31 & \bf 22.3$^*$ & 76.5 & 0.25 \\
\bf TransferTransfo+PE &  17.41 & 21.1 & \bf 82.0$^*$ & \bf 0.35 & 18.15 & 21.8 & \bf 77.2$^*$ & \bf 0.28$^*$ \\
\bf DeepCopy & 36.35 & 12.2 & 52.7 & 0.11 & 36.77  & 13.7& 49.6 & 0.07 \\
\bf GPMN & 36.11 & 13.5 & 54.9 & 0.15 & 36.45  & 14.8 & 51.4 & 0.10 \\ 
\hline
\bf TransferTransfo w/o persona & 19.87 & 18.4 & 67.3 & 0.04 &  - & - & - & - \\
\bf Persona Approximator & 18.33 & 19.8 & 73.3 & 0.22 & 18.59 & 20.4 & 71.2 & 0.16  \\ 
\bf Persona Generator & \bf 17.31$^*$ & 21.1 & 81.2 & 0.34 & \bf 18.11 & 21.9 & 76.8 & \bf 0.28$^*$ \\ 
\hline\hline
\end{tabular}}
\end{center}
\caption{\label{result1} Performance of dialogue models on automated evaluation metrics in the PersonaChat testset. ``Self persona'' means that the model is conditioned on the persona description of itself while ``their persona'' means the model is conditioned on the persona of its dialogue partner. We report the median as 5 random runs as the result. $^{*}$ denote statistically significant with p-value $<$ 0.05.}
\end{table*}

To explore to what extent our proposed approach is able to personalize dialogue agents, we compare two variants of our model which incorporate the persona approximator method and the persona generator method with the following baseline models:
\begin{itemize}
    \item \textbf{TransferTransfo}  A Transformer-based dialogue response generation pre-trained on general monolingual corpus by~\citet{TransferTransfo} and fine-tuned on Personachat by pre-pending all persona descriptions at the begining of the dialogue context.
    \item \textbf{TransferTransfo w/o persona} The same pre-trained TransferTransfo model fine-tuned on Personachat dataset without using persona information during training or inference.
    \item \textbf{TransferTransfo+PE} A transformer-based dialogue model based on pre-trained TransferTransfo model and fine-tuned by training a transformer-based persona encoder to provide persona embedding information.
    \item \textbf{DeepCopy} An RNN-based hierarchical
pointer network, which leverages copy mechanism to integrate persona~\cite{yavuz2019deepcopy}.
    \item \textbf{GPMN} Generative Profile Memory Network~\cite{zhang2018personalizing} is an RNN-based model that encodes persona as memory representations in a memory network. 

\end{itemize}
Both of our models (Persona Approximator and Persona Generator) are based on pre-trained TransferTransfo~\citep{TransferTransfo} and fine-tuned on Personachat. Specifically, the dialogue generation model is a 12-layer decoder-only transformer with masked self-attention heads (768-dimensional states and 12 attention heads). Fine-tuning hyperparameters are kept the same with~\citet{TransferTransfo}. To make the model compatible with the encoder-decoder architecture described in the method section, we consider the hidden state of the last token in the transformer model as the context embedding. For the persona encoder, we share all layers except the last layer in the multi-task setting. The RNN-based baselines are trained from scratch and we used their original architecture and training methods in the original paper.



\subsection{Experimental Results}

\paragraph{Results on PersonaChat}

We first present the experimental results on the PersonaChat dataset where persona description is available during training. In this scenario, the persona detection model is trained in the same domain as the response generation model.

The results of automated evaluation metrics are shown in Table 1. First, we can see that models explicitly incorporate textual persona descriptions, including the dialogue model that incorporate a persona encoder (i.e., \textbf{TransferTransfo+PE}) or pre-pend persona descriptions (i.e., \textbf{TransferTransfo}), outperform the baseline model that does not exploit persona information by a relatively large margin in all automated metrics. Also, dialogue models with a pre-trained Transformer model (i.e., TransferTransfo) substantially outperform RNN-based models that are trained from scratch.

As for our proposed approaches, we find that both persona detection models substantially improve the performance upon the baseline with the pre-trained TransferTransfo model without using persona information. It also significantly outperforms several models based on RNNs and use persona description during test time. When comparing the proposed two persona detection models, it is clear that the persona generator method performs much better than the persona approximator. More surprisingly, we find that it outperforms the competitive \textbf{TransferTransfo} model on several automated metrics despite not using any persona information at test time. We hypothesis that it is because the persona generator is trained with the reconstruction loss which is a useful supervision signal that is complementary to the MLE objective. In contrast, the persona encoder is trained jointly with the dialogue model by simply maximizing the likelihood of gold responses and may not actually capture the persona information.

When comparing the performance of our proposed approaches trained with either ``self persona'' and ``their persona'', we can see that training the persona detection to predict the persona information of the dialogue system itself helps the model to maintain a consistent persona, thus improving the consistency of generated responses. In contrast, training the persona detection model to predict the persona of its dialogue partner helps the model to generate more diverse responses.

\begin{table*}[t!]
\begin{center}
\resizebox{0.95\textwidth}{!}{
\begin{tabular}{ll|cccc}
\hline\hline
\bf Model & \bf Persona & \bf Fluency & \bf  Engagingness & \bf  Consistency & \bf  Persona Detection \\ \hline
\bf TransferTransfo & self &  3.49 &  3.47 &  3.47 & 0.85  \\
\bf TransferTransfo & their &  3.43 &  3.54 &  3.39 & 0.79  \\
\bf TransferTransfo & both &  3.55 &  3.63 &  3.51 & \bf 0.88  \\ \hline
\bf TransferTransfo+PE & self &  3.47 &  3.49 &  3.45 & 0.83  \\ 
\bf TransferTransfo+PE & their &  3.45 &  3.51 &  3.36 & 0.80  \\
\bf TransferTransfo+PE & both & 3.53 &  3.65 &  \bf 3.53 & 0.86  \\ \hline

\bf DeepCopy & self &  2.99 &  2.95 &  2.99 & 0.64 \\ 
\bf DeepCopy & their & 2.93 & 2.97 & 2.97 & 0.60 \\
\bf GPMN & self &  3.04 &  2.96 &  3.04 & 0.66 \\ 
\bf GPMN & their & 2.96 & 2.97 & 3.00 & 0.61 \\    \hline
\bf TransferTransfo w/o persona & $-$ & 3.28 & 3.13 & 3.17 & 0.62 \\
\bf Persona Approximator & self &  3.37 &  3.32 &  3.27 & 0.75 \\  
\bf Persona Approximator & their & 3.30 & 3.35 & 3.19 & 0.72  \\
\bf Persona Generator & self  &  3.50 &  3.51 &  3.43 & 0.85 \\
\bf Persona Generator & their & 3.45 &  3.59 & 3.31 & 0.80 \\
\bf Persona Generator & both & \bf 3.58$^{*}$ & \bf 3.67$^{*}$ & 3.47 & \bf 0.88 \\
\hline\hline
\end{tabular}}
\end{center}
\caption{\label{result1} Human evaluation of dialogue models with different personalization approaches on the PersonaChat dataset. $^{*}$ denote statistically significant with p-value $<$ 0.05.}
\label{human}
\end{table*}

\begin{table}[t!]
\begin{center}
\resizebox{\linewidth}{!}{
\begin{tabular}{ll|ccc}
\hline\hline
\bf Model & \bf Per & \bf Fluen & \bf  Engag & \bf  Consis \\ \hline
\bf Trans w/o persona & $-$ & 3.31 & 3.37 & 3.41 \\ \hline
\bf Persona Generator & self & \bf 3.50 & 3.48 & \bf 3.55 \\
\bf Persona Generator & their & 3.43 & \bf 3.55 & 3.51   \\ 
\hline\hline
\end{tabular}}
\end{center}
\caption{\label{result1} Performance of dialogue models with different personalization approaches on the Dailidialog dataset, persona encoder is not applicable as no persona description is available.}
\label{transfer}
\end{table}

Human evaluation results are shown in Table \ref{human}. We can see that dialogue models which explicitly incorporate textual persona descriptions significantly improves all human evaluation metrics.

As for our proposed approaches, we find that both proposed persona detection models can improve the consistency, engagingness, and persona detection accuracy upon the baseline seq2seq model without sacrificing the fluency of generated responses. The persona generator performs better than the persona approximator, which is consistent with the results in the automated evaluation. 
In addition, the persona generator model performs comparably and even better when compared with the competitive \textbf{TransferTransfo} baseline. This demonstrates that our proposed method can effectively personalize dialogue agents without relying on pre-defined persona descriptions at test time.

Similarly, we find that when conditioning on ``self persona'' as incorporating the persona description helps dialogue agents maintain a consistent profile throughout the conversation. Again, when conditioned on ``their persona'', the dialogue agent learns to predict the profile of its dialogue partner, which helps generate more engaging and personalized responses. Based on this motivation, we also conduct experiment with both ``their'' and ``self'' persona at the same time. We find this make significant future improvement and enabling dialogue agent to generate dialogue responses that are both engaging and consistent.

\paragraph{On the transferability of persona detection models}

As persona descriptions are not available in most scenarios and datasets, we aim to enable dialogue agent personalization for dialogue models trained in datasets where no persona description is available with a persona detection model pretrained on PersonaChat. To test the transferability of trained persona detection models, we combine persona detection models pretrained on the PersonaChat dataset with dialogue systems trained on the Dailydialog dataset. The pretrained persona detection models are fine-tuned jointly with the pretrained dialogue model by maximizing the likelihood of ground-truth responses. 
The results are shown in Table \ref{transfer}. We can see that transferring pre-trained persona detection models in the target dialogue domain is able to improve the performance of dialogue models. Specifically, predicting self-persona improves the consistency of the dialogue agent while detecting the persona of the dialogue partner improves the engagingness of generated responses. The experimental result also confirms the effectiveness of the proposed persona generator model and the persona reconstruction loss.

\subsection{Ablation Study}

To further understand the proposed models, we conduct an ablation study that focuses on: 1) the effectiveness of the multi-task learning architecture and the multi-task objective of persona detection models, and 2) the effect of available dialogue history length on the performance of persona detection models. We employ the dialogue response generation model with persona generator with self persona as the full model and compare it with the following ablated variants: (1) \textbf{first half:} The variant where only the first half of conversations are used as the test set, which makes the input dialogue history for persona generator shorter. (2) \textbf{second half:} The counterpart of \textbf{first half} where the available dialogue histories for persona generator are longer. (3) \textbf{w/o shared layers:} The variant where the persona generator does not share its first layer with the encoder of the dialogue model. (4) \textbf{w/o joint training:} The variant where the persona generator is exclusively trained with the reconstruction loss without jointly training with the MLE objective.

\begin{table}[t!]
\begin{center}
\resizebox{1.0\linewidth}{!}{
\begin{tabular}{lcccc}
\hline\hline
\textbf{Model} & \textbf{perplexity}  & \textbf{Dst} & \textbf{Hits@1} & \textbf{Cons} \\ \hline
\bf Trans w/o Persona  & 19.87 & 18.4 & 67.3 & 0.04 \\ 
~ - first half & 23.48 & 15.2 & 62.5 & -0.01 \\  
~ - second half & 17.16 & 21.3 & 71.3 & 0.05 \\\hline
\bf Persona Generator & 17.31 & 21.1 & 81.2 & 0.34 \\
~ - first half & 19.72 & 20.0 & 77.6 & 0.28 \\ 
~ - second half & 16.04 & 22.6 & 84.7 & 0.38 \\
~ - w/o shared layers & 18.67 & 20.6 & 80.1 & 0.30 \\
~ - w/o joint training & 18.55 & 20.4 & 80.5 & 0.31 \\

\hline \hline
\end{tabular}}
\end{center}
\caption{\label{result1} Results of the ablation study}
\label{ablation}
\end{table}

The results of the ablation study are shown in Table \ref{ablation}. We can see that both sharing layers and joint training improve the performance of the persona detection model, which demonstrates the effectiveness of multi-task learning in our task. As for the influence of the length of the dialogue history, we find that the proposed persona generator model performs better when giving longer dialogue history (i.e., the second half of the conversation), which is demonstrated by a larger relative improvement compared with the sequence-to-sequence baseline given the same dialogue history. This is reasonable as longer dialogue history may provide richer information and help detect persona better. It also suggests that our approaches may be more effective for dialogue agents that aim to conduct relatively long dialogues with humans. This problem is similar to the well-known cold-start problem in the field of recommend systems. However, this does not suggest that our proposed approach is not useful for most application scenarios where the dialogue agent must start the dialogue from scratch. In contrast, our model will continually track the persona information of both the dialogue agent itself and the dialogue partner, thus maintaining a consistent persona throughout the progress of the dialogue and gradually improve the engagingness of generated responses with the dialogue going on. In addition, the ability to automatically infer the persona information of the dialogue partner is also beneficial for real-world applications, where although we can pre-define a persona for the dialogue agent, the users' persona is not always available.

\subsection{Qualitative Analysis}

\begin{table}[t!]
\begin{center}
\resizebox{\linewidth}{!}{
\begin{tabular}{lc}
\hline\hline
\bf \text{No persona} &  I don't know what you could not do ?\\
\bf \text{Trans + PE w/ self}  & 
I am going to the club now. \\
\bf \text{Trans+ PE w/ their}  & 
Do you want to play frisbee or something? \\
\bf \text{PG w/ self}  & 
okay I am going to make a cake. \\
 - Generated Persona: &  ... I craving eating cake... \\
\bf \text{PG w/ their}  & 
I prefer that let's watch tv together. \\
 - Generated Persona: &  ... I like TV show... \\
\hline \hline
\end{tabular}}
\end{center}
\caption{\label{result1} Case study of the continuation of the conversation shown in Table 1 in the Appendix.}
\end{table}

To better understand the proposed method intuitively, we conduct a case study by feeding different variants of the dialogue model with the dialogue history presented in the Appendix and generate different continuations of the conversation. The next utterances generated by different model variants are shown in Table 6. We can see that the vanilla sequence-to-sequence dialogue model generates an irrelevant response that is not engaging. In contrast, both the persona encoder which takes the predefined persona description and the persona generator which infers the persona from dialogue history enables the dialogue agent to generate consistent and relevant responses, which are likely to be more engaging for the dialogue partner. In addition, we present the outputs of the decoder in the persona generator, which demonstrates that the proposed approach is more interpretable.  

\section{Conclusion}

In this paper, we propose a novel dialogue personalization approach that automatically infers the current speakers' persona based on the dialogue history, which enables neural dialogue systems to generate personalized dialogue responses without using persona description at test time. Our experiments on the PersonaChat dataset show that the proposed models can improve the model's consistency and engagingness when conditioning on the inferred persona information of the dialogue agent itself or the dialogue partner. We also conduct experiments on the Dailydialog dataset where persona description is not available and find that pre-trained persona detection models can be successfully transferred to other datasets without annotated persona descriptions. This further demonstrates the potential of our approach to enable personalized dialogue response generation for various domains where persona descriptions are not available or expensive to collect.



\bibliography{aaai21}
\bibliographystyle{aaai21}

\end{document}